\documentclass{ecai} 
\usepackage{latexsym}
\usepackage{times}
\usepackage{helvet}
\usepackage{courier}
\usepackage{natbib}
\usepackage{amsmath,amsfonts,amssymb}
\usepackage{algorithm}
\usepackage{algpseudocode}
\usepackage{csvsimple}
\usepackage{xcolor}
\usepackage{comment}
\usepackage{graphicx}
\usepackage{subcaption}
\usepackage{booktabs}

\newcommand{\name}{\text{\it name}}
\newcommand{\pre}{\text{\it precond}}
\newcommand{\effect}{\text{\it effect}}
\newcommand{\add}{\text{\it effect}^{+}}
\newcommand{\del}{\text{\it effect}^{-}}

\newcommand{\addposseffect}{\text{\it poss-effect}^{+}}
\newcommand{\delposseffect}{\text{\it poss-effect}^{-}}

\newcommand{\posseffectapprox}{\text{\it poss-effect}_{*}}
\newcommand{\addposseffectapprox}{\text{\it poss-effect}^{+}_{*}}
\newcommand{\delposseffectapprox}{\text{\it poss-effect}^{-}_{*}}

\newtheorem{definition}{Definition}

\begin{document}
 
 \begin{frontmatter}

\paperid{1160}

\title{SibylSat: Using SAT as an oracle to perform a greedy search on TOHTN Planning}

\author[A]{\fnms{Gaspard}~\snm{Quenard}\thanks{Corresponding Author. Email: gaspard.quenard@univ-grenoble-alpes.fr.}}
\author[A]{\fnms{Damier}~\snm{Pellier}}
\author[A]{\fnms{Humbert}~\snm{Fiorino}} 

\address[A]{ University Grenoble Alpes, LIG, F-38000 Grenoble, France}

\begin{abstract}
This paper presents SibylSat, a novel SAT-based method designed to efficiently solve totally-ordered HTN problems (TOHTN). In contrast to prevailing SAT-based HTN planners that employ a breadth-first search strategy, SibylSat adopts a greedy search approach, enabling it to identify promising decompositions for expansion. The selection process is facilitated by a heuristic derived from solving a relaxed problem, which is also expressed as a SAT problem. Our experimental evaluations demonstrate that SibylSat outperforms existing SAT-based TOHTN approaches in terms of both runtime and plan quality on most of the IPC benchmarks, while also solving a larger number of problems.
\end{abstract}

\end{frontmatter}

\section{Introduction}

Hierarchical Task Network (HTN) planning \cite{erol1994umcp} is a field of Artificial Intelligence dedicated to decomposing complex tasks into simpler subtasks. This hierarchical approach improves planning speed and scalability, especially in practical applications \cite{georgievski2015htn}. Unlike classical planning, HTN planning introduces abstract tasks representing complex, non-executable actions, and decomposition methods that break down abstract tasks into partially ordered sets of primitive actions and abstract tasks. The main goal of an HTN planner is to determine how an initial abstract task can be decomposed into a plan i.e., an executable sequence of primitive tasks.

In this paper, we focus our investigation on a highly popular subclass of HTN problems where the decomposition methods specify a totally-ordered list of primitive actions and abstract tasks to be executed in order to achieve an abstract task, namely Totally-Ordered HTN (TOHTN) planning. The prevalent approaches to TOHTN planning are either to use heuristic search, e.g., \cite{holler2018generic, holler2019guiding, holler2020htn}, to encode planning problems into STRIPS problems in order to benefit from the constant improvements of the classical planners, e.g.,  \cite{alford2009translating, alford2016bound, behnke2022making}, or to encode them into propositional logic, e.g., \cite{behnke2018totsat, schreiber2019tree, behnke2021block, schreiber2021lilotane} and use highly optimized solvers. The latter has attracted renewed interest over the last few years, especially with planners performing particularly well in recent International Planning Competitions (IPC). The work presented in this paper aligns with this trend.

In the realm of TOHTN planning, the prevailing strategy among current SAT planners is to employ a "breadth-first search" approach, bounded by the number of decompositions of the initial abstract task. Specifically, for a given bound $k$, the logical encoding considers plans where the initial task can be decomposed at most $k$ times. If no solution is found within this bound, $k$ is increased, broadening the solution search to include more intricate decompositions. This iterative process continues until a satisfactory solution is discovered or until a predefined maximum limit for $k$ is reached.

While this breadth-first search technique ensures completeness, guaranteeing that a solution, if it exists, will eventually be found, it operates as a blind search and can be particularly inefficient when dealing with large search spaces or complex problem instances. Unlike many planning techniques that utilize heuristic information to guide the solution search towards promising areas of the search space, the current SAT-based approaches to TOHTN planning do not build on heuristic search.

In this paper, we introduce SibylSat, a novel SAT-based approach that employs a greedy best-first search strategy. This strategy involves selectively expanding search areas based on a heuristic derived from solving a relaxed version of the planning problem, allowing the identification of promising decompositions. Unlike existing SAT-based TOHTN planners, SibylSat does not automatically expand all the pending decompositions when no solution is found. Instead, it intelligently determines which decompositions should be developed further based on this heuristic. Our approach considerably reduces the search space and lays the groundwork for the development of new heuristics and techniques to solve HTN problems by SAT encodings. Experimentally, we demonstrate that our greedy search approach, coupled with a heuristic function, improves performance on most IPC benchmarks compared to state-of-the-art SAT TOHTN approaches.

The article unfolds as follows: first, we introduce the concept of TOHTN planning. Next, we describe our planner, SibylSat. Finally, we compare SibylSat with other SAT-based TOHTN planners.

\section{HTN Planning Problem}

This paper adopts a description of TOHTN planning inspired by the formulation presented in \cite{behnke2018totsat, behnke2021block}, which is an adaptation of the HTN planning description outlined in \cite{geier2011decidability} tailored to totally ordered planning.

\subsection{Task, Action, Methods, and Task Networks}

The central concept in HTN Planning is the notion of \textit{tasks}. A \textit{task} is characterized by a name and a list of parameters. There are two types of tasks: \textit{primitive tasks} and \textit{abstract tasks}. While primitive tasks directly affect the state of the world, abstract tasks do not; however, they must be decomposed into primitive tasks via \textit{decomposition methods} before they can be executed. The HTN planning problem specifies how a method can decompose an abstract task into a \textit{task network}, which is a sequence of tasks (either abstract or primitive). A \textit{primitive task network} refers to a task network comprising solely primitive tasks.

A primitive task $a$ is analogous to an action in classical planning and is defined by a tuple $(\name(a), \pre(a), \effect(a))$. Here, $\name(a)$ represents the name of $a$, while $\pre(a)$ and $\effect(a)$ represent the sets of propositions for preconditions and effects, respectively. An action $a$ is considered executable in a state $s$, defined as a set of propositions describing the world, if and only if $\pre(a)$ is a subset of $s$. If $O$ represents the set of actions and $S$ represents the set of states, then the state transition function $\gamma: S \times O \rightarrow S$ is defined as follows: If $a$ is executable in $s$, then $\gamma(s, a) = (s \setminus \del(a)) \cup \add(a))$; otherwise, $\gamma(s, a)$ is undefined. The extension of $\gamma$ to action sequences, denoted $\gamma^*: S \times O^* \rightarrow S$, is defined straightforwardly.

A method $m$ indicates how an abstract task can be refined into a task network. It is defined as a tuple $(\name(m), c, w_m)$ where $c$ is an abstract task and $w_m$ is a task network. We refer to $c$ as the task of $m$ and $w_m$ as the subtasks of $m$. Given a task network $w = w_1 c w_2$ where $c$ is an abstract task, applying a method $m = (\name(m), c, w_m)$ to $w$ will result in the task network $w' = w_1 w_m w_2$ (written as $w  \rightarrow^{m} w'$). We write $w  \rightarrow^{*} w'$ if any sequence of decomposition methods exist which transforms $w$ into $w'$. For notation purposes, we define as $M(c) = \{m = (name(m),c,w_m) \ | \ m \in M\}$ the set of all the methods which can be applied to the abstract task $c$.

\subsection{Planning Problem and Solution}

We define a TOTHN planning problem as follows: 

\begin{definition} [TOHTN Planning problem]

A totally ordered HTN Planning problem $P$ is a tuple $(L, C , O ,M , c_I, s_I, g)$ where: $L$ is a finite set of propositions; $C$ is a finite set of abstract (or \textit{compound}) tasks; $O$ is a finite set of primitive tasks (or \textit{actions}); $M$ is a finite set of decomposition methods (or \textit{methods}); $c_I \in C$ is the initial abstract task; $s_I \subseteq L$ is the initial state and $g$ is the (possibly empty) goal state.
\end{definition}

Solving a planning problem $P$ involves finding a primitive task network (or \textit{plan}) $\pi$ which can be decomposed from the initial abstract task ($c_I \rightarrow^{\ast} \pi$) such that $\pi$ is executable in the initial state $s_I$ and reaches the goal $g$ after execution, i.e., $g \subseteq \gamma(s_I,\pi)$. Unlike classical planning, a solution to a TOHTN planning problem is not only a sequence of primitive tasks executable in the initial state of the problem and reaching a goal state, but it must also express the decomposition methods used which lead to this sequence of primitive tasks. Such a solution can be represented by a tree called a \textit{decomposition tree} \cite{geier2011decidability} which shows the full trace indicating how an abstract task is refined into a task network.

\begin{definition} [Decomposition tree]

A decomposition tree (DT) for a task network $w$ for a problem $P = (L, C , O ,M , c_I, s_I, g)$ is a tree $T = (N, E)$ with:

        \begin{itemize}
            \item $N$ - A set of nodes labelled by either a primitive task, an abstract task or a method,
            \item $E: N \rightarrow N^{*}$ - the edge function which provides for every node an ordered list of children $\langle e_1, e_2, ..., e_k \rangle$,
            \item For every inner node $n$ labelled by an abstract task $c$, $|E(n)| = 1$ and $E(n) = \langle n' \rangle$ where $n'$ is labelled by a method $m \in M(c)$.
            \item For every inner node $n$ labelled by a method $m$, let us consider $\langle t_1, t_2, \ldots, t_k \rangle$ the subtasks of $m$, then $|E(n)| = k$ and $e_i$ is labelled by the task $t_i$ for all $e_i \in E(n) = \langle e_1,e_2,\ldots, e_k\rangle$.
            \item For the sequence of leafs $L=\langle n_1^l,n_2^l, \ldots, n_k^l \rangle$, it holds that each leaf if labelled by either a primitive task or an abstract task and if we consider the corresponding sequence of tasks $\langle t_1, t_2, \ldots, t_k \rangle$, then we have $w = \langle t_1, t_2, \ldots, t_k \rangle$.
        \end{itemize}

\end{definition}

We can now formally define a decomposition tree solution as follows:

\begin{definition} [Decomposition Tree solution]
    Let $P = (L, C , O ,M , c_I, s_I, g)$ be a planning problem. Consider $T_{sol}$ as the decomposition tree for the task network $\pi$ for the problem $P$. The decomposition tree $T_{sol}$ is a solution for $P$ if and only if
    it achieves the following characteristics:

    \begin{enumerate}
        \item The root of $T_{sol}$ is the initial abstract task $c_I$.
        \item $\pi$ only contains primitive tasks.
        \item $\pi$ is executable in the initial state $s_I$.
        \item $\pi$ reaches the goal $g$ after execution.
    \end{enumerate}
\end{definition}

\section{Path Decomposition Tree and SAT planners}

Since finding a solution for an HTN planning problem is equivalent to finding a decomposition tree (DT) that satisfies some characteristics, 
one way to find a solution is to test all possible DTs that exist for a problem $P$. However, the search space representing all possible DTs is very large for most domains and infinite for recursive domains (domains in which an abstract task can be obtained by decompositions from the same abstract task). To address this issue, \cite{behnke2018totsat, schreiber2019tree} suggested creating structures that represent a subset of all possible DTs to check if they contain valid solutions. If no solution is found, these structures can be expanded to encompass additional DTs. These proposed structures are designed to ensure they can be expanded to include any possible DT for a given problem. Here, we introduce an isomorphic equivalent of their structures:

\begin{definition} [Path Decomposition Tree]
        A path decomposition tree (PDT) for a problem $P = (L, C , O ,M , c_I, s_I, g)$ is a tree $\Gamma = (N, E)$ with:

        \begin{itemize}
            \item $N$ - A set of nodes labelled by either a primitive task, an abstract task or a method,
            \item $E: N \rightarrow N^{*}$ - the edge function which provides for every node an ordered list of children $\langle e_1, e_2, ..., e_k \rangle$. \\

            We denote the root node of the PDT as $r_{\Gamma}$. We call the PDT well-formed if and only if:
            \item The root node of the PDT $r_{\Gamma}$ is labelled by the initial abstract task of the problem $c_I$.
            \item For every inner node $n$ labelled by an abstract task $c$, $|E(n)| = |M(c)|$ and $\forall m_i \in M(c), \exists e_i \in E(n)$ such that $e_i$ is labelled by the method $m_i$.
            \item For every inner node $n$ labelled by a method $m$, let us consider $\langle t_1, t_2, \ldots, t_k \rangle$ the subtasks of $m$, then $|E(n)| = k$ and $e_i$ is labelled by the task $t_i$ for all $e_i \in E(n) = \langle e_1,e_2,\ldots, e_k\rangle$.
            \item For every leaf $n^l$, $n^l$ is labelled by either a primitive task or an abstract task.
        \end{itemize}        
\end{definition}

\begin{figure}[ht]
    \centering
    \includegraphics[width=0.5\textwidth]{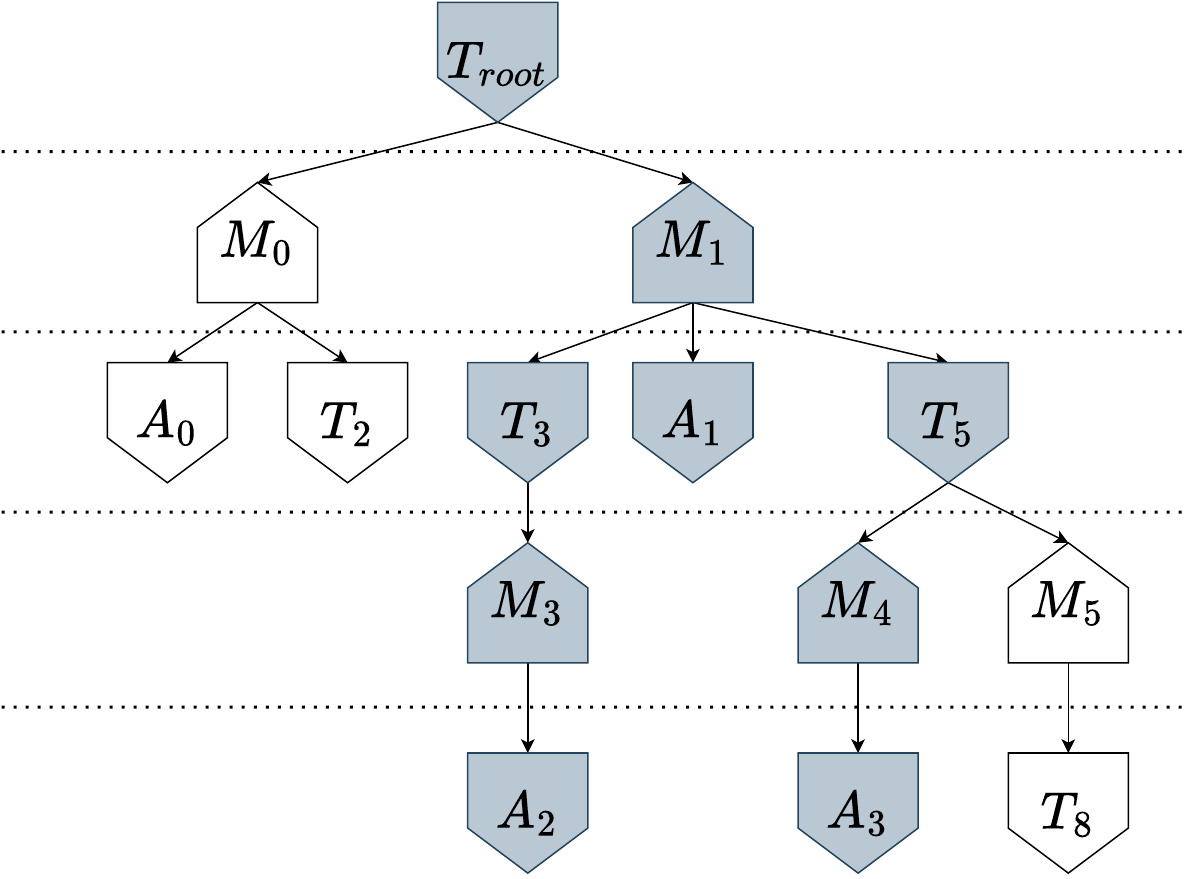}
    \caption{Example showing a PDT at some level of decomposition for a problem $P=(L, C , O ,M , c_I, s_I, g)$, where $T_i \in C, M_i \in M \text{ and } A_i \in O$. We see that this PDT does not contain all the DTs because the abstract tasks $T_2$ and $T_8$ are undeveloped. A potential solution DT is highlighted in grey.}
    \label{fig:example_and_or_tree_network_decomposition}
\end{figure}

Figure \ref{fig:example_and_or_tree_network_decomposition} illustrates a PDT for a problem $P = (L, C , O ,M , c_I, s_I, g)$ at some level of decomposition, and highlights a DT for the task network $\langle A_2,A_1,A_3 \rangle$ as a subtree of this PDT. This DT is a solution for $P$ if and only if $\pi = \langle A_2,A_1,A_3 \rangle$ is executable in $s_I$ and $g \subseteq \gamma(s_I,\pi)$. \\

\vspace{0.5cm} 

Searching for a solution DT in a PDT is the current approach used by all SAT-based TOHTN planners \cite{schreiber2019tree, schreiber2021lilotane, behnke2018totsat, behnke2021block}. The core idea is to encode a PDT (or an isomorphic equivalent data structure) into a formula such that this formula is satisfiable if and only if a solution DT exists in the PDT. If the formula is satisfiable, the solution DT can then be extracted from the satisfying assignment. If no solution is found, the PDT is expanded. All currently existing encodings develop the PDT by expanding all its abstract task leaf nodes (or \textit{pending nodes}). If the initial PDT is initialized with only a node representing the initial abstract task, \cite{behnke2018totsat} have shown that, by following this expansion approach, the $j^{th}$ PDT will capture all possible DTs with a maximum depth of decomposition of $j$. Therefore, this approach represents a "breadth-first search" along the depth of decomposition and is, as a consequence, complete.

\section{SibylSat planner}

\subsection{Planning approach}

Like other SAT-based TOHTN planners, SibylSat employs a PDT as a search space and adheres to a standard procedure of alternating between expanding the search space, encoding, and invoking a SAT solver to find a solution plan. However, a significant departure lies in the expansion strategy, which does not follow a breadth-first approach. Instead, SibylSat selects promising abstract tasks for expansion. 

The procedures are illustrated as a pseudo-code in Algorithm~\ref{algo:greedy_sibylsat}. The algorithm begins by initializing the PDT with a single node representing the initial abstract task of the problem. Then, the process alternates between two phases until a solution is found: one where SibylSAT seeks a solution DT within the PDT (which acts as the termination condition), and another where it expands the PDT by exploring first the most promising abstracts tasks. To achieve this, it focuses on finding and developing a \textit{promising} DT within the PDT, where 'promising' refers to a DT whose expansion (i.e., the development of all its pending nodes) could lead to a solution DT. The choice of the promising DT to develop is made by searching for a solution DT in a relaxation of the PDT where abstract task leaves are considered as actions. The DT solution found in this relaxed PDT is used as a heuristic to identify which pending nodes need to be developed in the PDT to reach a solution. Specifically, any abstract task that is part of the plan of the relaxed solution DT will have its corresponding leaf in the PDT developed. The search for a solution DT in the PDT, as well as a promising DT in its relaxed counterpart is done by a SAT solver.

A necessary (but not sufficient) condition to ensure the completeness of our planner is that any subtree of a solution DT must also be recognized as a promising DT in a relaxed PDT. Failure to meet this condition could result in overlooking viable solutions. Therefore, when relaxing the PDT, the preconditions and effects assigned to an abstract task must be carefully inferred to comply with this requirement. The approach for deducing them will be detailed in an upcoming section. Note that this implication is not bidirectional: a promising DT does not necessarily lead to a solution DT when further developed. The accuracy with which we can represent abstract tasks as actions (i.e., how well we can infer their preconditions and effects) brings us closer to achieving this equivalence. 
It should also be noted that a relaxed PDT typically contains several promising DTs available for further development. We do not use a deterministic method to select which promising DT to develop. Instead, we entrust this decision to our SAT solver, which acts as an oracle in this context. Specifically, we use an \textit{incremental SAT solver} (i.e., a solver that can be queried multiple times with a growing set of clauses and is able to preserve its knowledge base from previous solving iterations), which aids in maintaining a consistent trajectory when choosing a promising DT to develop as it incrementally builds upon previous decisions.  \\

Algorithm \ref{algo:greedy_sibylsat} is "greedy" because it expands the PDT by using the first relaxed solution DT it finds in its expansion phase, even if there are multiple relaxed solution DTs in a PDT. This method is complete for non-recursive planning domains because there is a limited number of DTs, and our algorithm eventually finds them all as it keeps expanding the PDT. However, this approach is generally non-terminating for recursive domains because the algorithm can keep expanding DTs in the PDT that cannot lead to a solution DT. We will explain how we can transform the algorithm \ref{algo:greedy_sibylsat} to be terminal in a later section.

\begin{algorithm}
\caption{SibylSat Planner}
\label{algo:greedy_sibylsat}
\begin{algorithmic}[1]
\Procedure{SibylSat}{$P=(L, C , O ,M , c_I, s_I, g)$}
    \State $PDT \gets \Call{InitializePDT}{P}$
    \State \Return $\Call{Greedy}{PDT}$
\EndProcedure
\Procedure{Greedy}{$PDT$}
    \State $DT_{sol} \gets \Call{FindSolution}{PDT}$
    \If{$DT_{sol} \neq \emptyset$}
        \State \Return $DT_{sol}$
    \EndIf

    \State $DT_{relaxed} \gets \Call{FindPromisingDT}{PDT}$
    \State $PDT \gets \Call{ExpandPDT}{PDT, DT_{relaxed}}$ 
    \State \Return \Call{Greedy}{$PDT$}
    
\EndProcedure

\end{algorithmic}
\end{algorithm}

\subsection{Example}

To illustrate our algorithm, we detail how the PDT of a problem $P$ is developed by our planner in Figure \ref{fig:example_and_or_tree_network_decomposition}. SibylSat is initialized by creating a PDT with a single root node, $T_{root}$, representing the problem's initial abstract task. It then alternates between a search phase and an expansion phase until a solution DT is found. \\

In the initial search phase, since $T_{root}$ is abstract, the PDT does not contain a solution DT, prompting the first expansion phase. In the expansion phase, SibylSat uses the relaxation technique
to transform all the abstract task leaves, in this case, $T_{root}$, into actions. A search is then performed within the relaxed PDT, which identifies the relaxed DT of the plan 
$\langle T_{root} \rangle$
as a solution. This DT is then developed within the PDT, leading to an expanded PDT encapsulating the initial three layers of Figure \ref{fig:example_and_or_tree_network_decomposition}, before returning to the search phase. \\

The second search phase fails again to find a solution DT in the PDT, leading to another expansion phase. In this phase, the PDT is relaxed once more by treating the abstract task leaves $T_2$, $T_3$ and $T_5$ as actions and trying to find a solution DT in this relaxed PDT. Notably, within this relaxed PDT, two potential solution DTs emerge: one involving the task network $\langle A_0, T_2 \rangle$ and another comprising $\langle T_3, A_1, T_5 \rangle$. 
Assuming that these task networks are both executable in the problem's initial state and can reach the goal after execution, the SAT solver selects one of them for further development. 
In this example, the SAT solver chooses the DT of the task network $\langle T_3, A_1, T_5 \rangle$, prompting SibylSat to expand the corresponding nodes $T_3$ and $T_5$, which leads to the whole PDT in Figure \ref{fig:example_and_or_tree_network_decomposition} before returning to the search phase. \\

In this search phase in the PDT, SibylSat identifies the solution DT with the plan $\langle A_2, A_1, A_3 \rangle$, leading to a successful termination of the planner. However, should this solution prove non-executable in the initial state, the absence of other possible solution DTs would have led to another expansion phase of the PDT. In that case, if both the DT of the task networks 
$\langle A_0, T_2 \rangle$ and $\langle A_2, A_1, T_8 \rangle$ 
are relaxed solution DTs, then, the SAT solver would have returned either of those two, which would have led to developing either the node $T_2$ or the node $T_8$.

\section{Searching for solution DT in PDT}\label{sec:explore_pdt}

During the search phase, our planner needs to determine whether a PDT contains a solution DT as its subtree. To achieve this, the approach of SAT-based TOHTN planners is to create a SAT formula that is satisfiable if and only if a solution DT exists in the PDT. Given that our search space is structurally congruent with those proposed by \cite{schreiber2021lilotane, schreiber2019tree, behnke2018totsat, behnke2021block}, we can use any of the encodings introduced in these papers to search for a solution DT. We opted for the encoding used by the Lilotane planner specifically because it can be employed with an incremental solver, which enables our solver to keep knowledge of previous solving iterations when looking for solution DTs and relaxed solution DTs. Given that the encoding is no different from that proposed by Lilotane for a PDT, we do not detail the specific SAT rules used to search for a solution DT in this paper.

\section{Expansion of the PDT}

In this section, we present the expansion phase of the PDT. This phase occurs when there are no solution DTs in the current PDT (i.e., the search procedure has failed to find a solution DT). As such, some pending leaves need to be expanded to allow the PDT to capture a larger set of DTs. The core idea of our expansion phase is to find a DT that is a subtree of a solution DT. If such a DT could be found, developing it would lead to expanding the PDT toward having a solution DT as its subtree. However, finding a DT leading to an executable primitive plan is not feasible, as it would require knowing a solution DT for the problem in the first place. As such, we are interested in an approximation. This approximation takes the form of finding instead a promising DT: a DT that may lead to a solution DT when developed. We define a DT as 'promising' if its relaxed task network is executable in the initial state and reaches the goal after execution. To relax a task network, we convert all its abstract tasks into actions with preconditions and effects based on the possible decompositions of the respective abstract task. As such, finding a promising DT is equivalent to finding a solution DT in a relaxed PDT where all abstract task leaf nodes are converted into actions.  

We now turn to the process by which the preconditions and effects of an abstract task are inferred. As detailed in the planning approach, a necessary condition for completeness is that all subtrees of a solution DT are recognized as promising DTs. Consequently, the preconditions and effects inferred for an abstract task must be applicable for all refinements into primitive task networks of the respective abstract task.

First, we explain the inference of preconditions for an abstract task. These preconditions must encapsulate the facts necessary for any feasible refinement of an abstract task. \cite{olz2021revealing} have referred to such preconditions as 'mandatory preconditions', a concept used in several planners \cite{schreiber2021lilotane, magnaguagno2020hypertension} to prune DTs in their search space that do not lead to an executable sequence of primitive tasks. In our planner, we employ the algorithm for inferring mandatory preconditions proposed by \cite{schreiber2021lilotane}.

Let us now consider the effects of an abstract task. Unlike actions, where effects are explicitly defined, abstract tasks introduce uncertainty in the resulting state. Indeed, for an action $a$, the transition $\gamma(s,a) = (s \setminus \del(a)) \cup \add(a)$ precisely defines the post-execution state. In contrast, it is difficult to properly define the post-execution state of an abstract task since it can be decomposed into multiple different primitive task networks with different post-execution effects. Our solution for this problem is to compute all the $\addposseffect(t)$ (respectively $\delposseffect(t)$) of an abstract task, which correspond to all the positive (respectively negative) facts that may be caused by a refinement of the abstract task.

If we successfully compute these two sets, we can then define an overestimation of the post-execution state of an abstract task $t$. This is achieved by considering the execution of $t$ as leading to a nondeterministic state, which falls within a range determined by subtracting any subset of $\delposseffect(t)$ and adding any subset of $\addposseffect(t)$ to the pre-execution state.

\cite{behnke2021block} have shown that computing the exact set of possible effects of an abstract task is PSPACE or EXPTIME-complete depending on the hierarchical decomposition structure of the respective task. However, it is possible to find an over-approximation of them in polynomial time. For our algorithm, we compute an over-approximation of the possible effects of an abstract task (denoted $\posseffectapprox$) with the following formula: 

\[\addposseffectapprox(t) =  \bigcup_{t' \in subtasks(m), m \in M(t)} \addposseffectapprox(t')\]
\[\delposseffectapprox(t) =  \bigcup_{t' \in subtasks(m), m \in M(t)} \delposseffectapprox(t')\]

If $t$ is an action, then: 

\[\addposseffectapprox(t) = \add(t)  \]
\[\delposseffectapprox(t) = \del(t)  \]

If $t$ is recursive, we avoid infinite recursion by removing $t$ from all the possible subtasks of all the methods. \\

Note that the nondeterministic post-execution state of an abstract task, as defined above, is often an over-approximation of the post-execution state that one specific concrete refinement of an abstract task can actually produce, due to it encompassing effects from various refinements. Although \cite{olz2023can} proposed a method to generalize the possible effects of an abstract task by differentiating them based on the outcomes of each refinement, this method is computationally prohibitive due to its high cost. Nevertheless, it is still possible to refine the nondeterministic post-execution state of an abstract task computed by eliminating inconsistencies that arise when combining effects from different refinements.

\begin{figure}[h]
    \centering
    \includegraphics[width=0.5\textwidth]{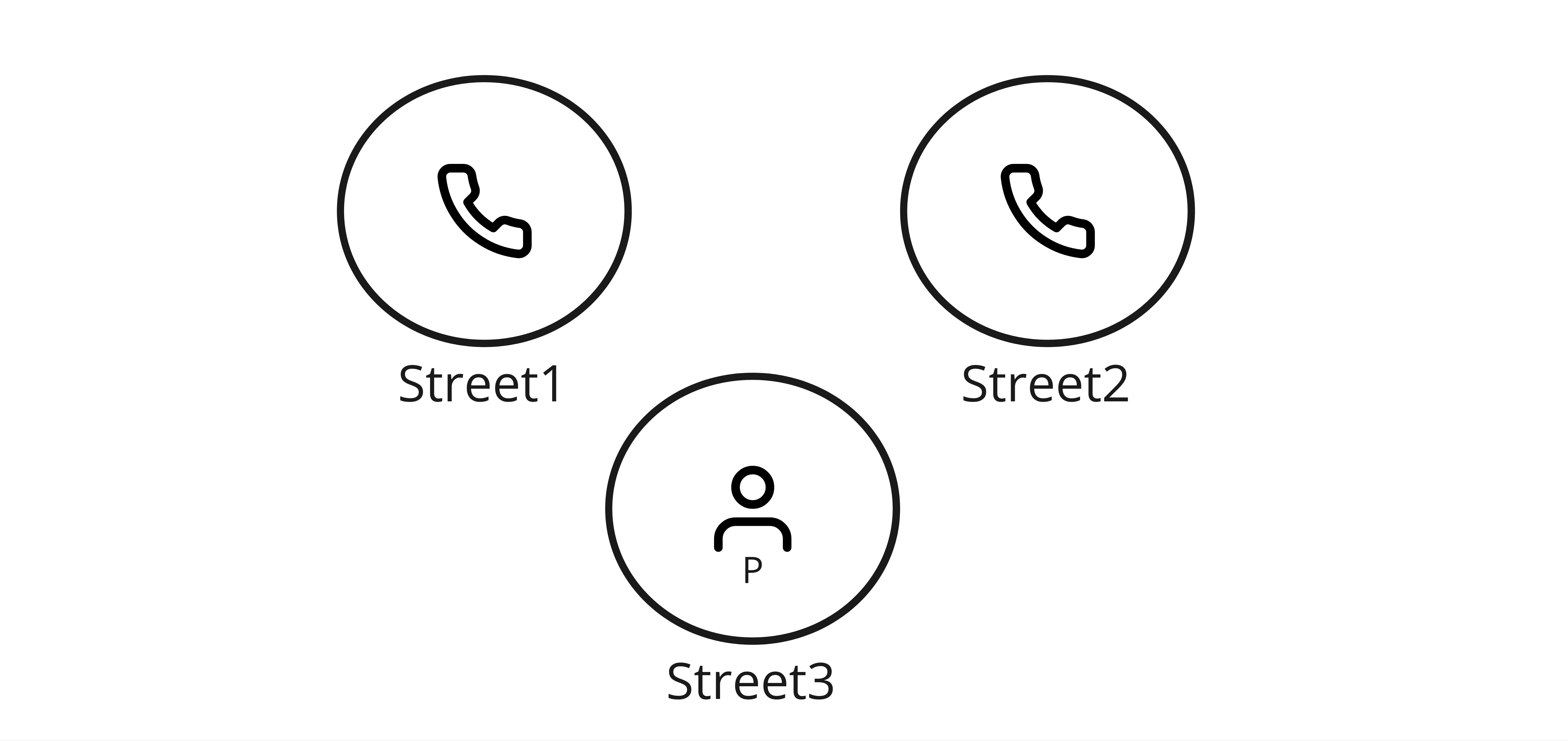}
    \caption{Toy world state.}
    \label{fig:example_mutex}
\end{figure}

\vspace{0.5cm} 

Consider the toy example illustrated in Figure \ref{fig:example_mutex}. The figure shows a world state with three locations: Street1, Street2, and Street3. A person, denoted as 'P', is located at Street3 and needs to get a taxi. Consider an abstract task $call\_taxi()$, meant to accomplish this by first reaching a location with a telephone booth and then calling a taxi. Since in this example, there is a telephone booth in both Street1 and Street2, the possible effects of the abstract task $call\_taxi()$ will have the predicates $at(P, Street1)$ and $at(P, Street2)$. If this task is treated as a direct action, using our formula to compute its post-execution state might create an inconsistency where the person is at both Street1 and Street2 simultaneously. Such an incoherent state of the world could result in finding relaxed solution DTs that are impractical - meaning they cannot yield a valid solution DT when further developed. This often leads to developing nodes that turn out to be useless for finding a valid solution DT.

Our approach to reducing these types of inconsistencies in the post-execution state of an abstract task involves the utilization of mutexes. Mutexes define the set of facts that cannot coexist within the same reachable state. In our example, the predicates $at(P, Street1)$ and $at(P, Street2)$ are mutex because a person cannot be at two places at the same time. To infer those mutexes, we use the  algorithm proposed by \cite{fivser2020lifted} which infers mutex groups: groups of predicates where each predicate in a group is mutex with all the other predicates in the same group. Using those mutexes, we can improve the quality of the post-execution state of an abstract task by preventing incompatible possible effects from occurring.

In practice, this is achieved by applying an 'at-most-one' constraint in our SAT encoding to prevent mutually exclusive effects of an abstract task from occurring simultaneously. There are various methods to encode these 'at-most-one' constraints in SAT, as discussed by \cite{nguyen2020empirical}. 
For our empirical studies, we tried three 'at-most-one' encodings: classical pairwise, Binary, and Bimander (using (m = $n/2$) and (m = $\sqrt{n}$)). Our results show that, using the solver Glucose \cite{audemard2018glucose}, the classical pairwise encoding seems to perform marginally better than the other tested encodings, albeit with a larger memory footprint.
Our experiments have demonstrated that in many domains, employing mutexes results in finding higher quality abstract solution DTs and, consequently, developing fewer nodes in the PDT before finding a solution DT.

\subsection{Searching for solution in relaxed PDT}

Once the PDT has been relaxed to replace all pending abstract tasks with their corresponding actions, the search process mirrors that of the non-relaxed PDT, with the notable distinction that the effects of the abstract tasks are nondeterministic. This uncertainty means that an effect listed in $\posseffectapprox(c)$ might not necessarily occur even if $c$ is executed. Consequently, when employing a SAT encoding to search for a solution DT in a relaxed PDT, the effects of an abstract task leaf node cannot be encoded in the same manner as the effects of an action, which are represented as $action \implies eff$. Instead, for each possible effect of an abstract task leaf node, we must encode the fact that if the abstract task leaf node is in a solution DT, then each of its effects \textit{may} happen in its post-execution state.
 Lilotane already has rules to encode such uncertainty in the post-execution state of an abstract task, called \textit{frame axioms}. Frame axioms are rules to ensure that if a predicate value changes, then a specific abstract task or action that has this predicate in its effects must have been executed. By using frame axioms to encode the possible effects of an abstract task, we correctly take into account their nondeterministic effects.

\subsection{Expanding PDT leafs}

Finally, let us explain how we develop the pending nodes of a PDT based on a solution DT found in a relaxed PDT. In the relaxed PDT, all abstract task leaf nodes are considered actions. Consequently, the plan $\pi$ in the solution DT may include abstract tasks. For each of these abstract tasks, we develop the corresponding node in the PDT, as well as all the immediate child method nodes. This approach ensures that all leaves of the PDT contain either primitive or abstract tasks. For example, as illustrated in Figure \ref{fig:example_and_or_tree_network_decomposition}, if the plan in the solution DT for the relaxed PDT is $\langle A_0,T_2 \rangle$, then the node containing $T_2$ and its immediate method children will be developed.

Note that in our implementation, we use the search space proposed by the Lilotane planner, where each node contains all the abstract tasks that may occur simultaneously at some degree of refinement. To adhere to the incremental encoding provided by Lilotane, where some rules depend on the full set of operations within a node, we develop all the abstract tasks within the same node when such a node is expanded. A potential next contribution could be to alter the encoding to allow for the selective development of only the abstract tasks found in the relaxed solution DT.

\section{Ensuring completeness for recursive domains}

As the current description given for our planner is greedy instead of the "breadth-first search" used by the other TOHTN SAT planners, this approach may suffer from non-termination for recursive domains.

A possible approach to mitigate this issue would be to limit the maximum decomposition depth of the PDT with an iterative deepening approach, ensuring that all DTs with a decomposition depth of up to $k$ are explored before incrementally increasing the bound. In practice guessing the initial depth of decomposition and how to increment it can be challenging, as it depends on the domain's hierarchy structure as well as the problem's characteristics (objects, initial abstract task and initial state).

Our approach adopts a more nuanced technique that relies directly on the domain's structure. We observe that for each non-recursive abstract task, the process of repeatedly decomposing this abstract task is finite. Therefore, we only need to limit the decomposition of recursive abstract tasks to ensure completeness. We use the following approach: for each recursive abstract task $t$ in the PDT, we prevent any of its children or transitive children from being identical to $t$. Specifically, if a node labeled by a method $m$ is a child or transitive child of a node labeled by an abstract task $t$, and one of the method $m$'s subtasks is $t$, then we exclude $m$ from the PDT. This technique ensures that the exhaustive expansion of our PDT is finite. If no solution DT is found when the PDT is fully developed with these restrictions, we then incorporate all the omitted methods $m$ and their subtasks into the PDT and repeat the process. This approach allows our algorithm to achieve completeness. \cite{magnaguagno2020hypertension} have proposed a very similar technique to avoid infinite loops for their TOHTN planner HyperTensionN.

Empirically, this methodology has been effective. Our planner was capable of solving problems from both the 2020 and 2023 benchmarks of the International Planning Competition (IPC) without needing to insert even once the missing methods into the PDT.

\section{Evaluation}

\subsection{Planners}

We compared SibylSat against other state-of-the-art SAT-based TOHTN planners to evaluate its performance. The evaluation involved three planners: SibylSat\footnote{source code is available at https://github.com/gaspard-quenard/sibylsat}, our planner with the approach to ensure completeness linked with the Glucose solver \cite{audemard2018glucose}; Lilotane \cite{schreiber2021lilotane}, a Lifted SAT TOHTN planner and runner-up of the 2020 International Planning Competition (IPC); and pandaPIsatt-1iB \cite{behnke2021block}, an improved version of their original TOHTN planner totSAT \cite{behnke2018totsat}. The experiments were conducted on a system with an Intel Core i7-12700H CPU and 32GB of RAM. Each problem instance was given a maximum runtime of 10 minutes. The evaluations covered all TOHTN benchmarks proposed in the IPC 2020 and IPC 2023.

\subsection{Evaluation Metrics}

The performance of the planners is evaluated using two different metrics:

\subsubsection{IPC Score}
 The IPC score measures the speed of the planner in solving a problem. The score ranges from 0 to 1 for each problem, with 1 indicating that the planner solved the problem in less than one second and 0 indicating a timeout. The IPC score is calculated as follows:

 \[
\text{IPC Score} = 
\begin{cases} 
0 & \text{If no plan is found} \\
\min\left(1, 1 - \frac{\log(t)}{\log(T)}\right) & \text{if a plan is found.}
\end{cases}
\]

where $T$ is the maximum runtime allowed for finding a plan, and $t$ is the time taken to find a plan, in seconds.

\subsubsection{Quality Score}

The quality score measures the makespan (i.e., the total number of actions in a plan). It ranges from 0 to 1, with 1 indicating the shortest plan and 0 indicating that no solution was found. The quality score is calculated as follows:

 \[
\text{Quality Score} = 
\begin{cases} 
0 & \text{If no plan is found} \\
\frac{C^{ref}}{C} & \text{if a plan is found.}
\end{cases}
\]

where $C$ is the makespan of the plan found by the planner and $C^{ref}$ is the best makespan among all evaluated planners.
\subsection{Results}

\begin{table}[ht]
\begin{tabular}{lrlll}
\toprule
{Domain}& \rotatebox{90}{\# instances}& \rotatebox{90}{SibylSat}& \rotatebox{90}{Lilotane}& \rotatebox{90}{PandaPIsatt-1iB} \\
\midrule
\hline
AssemblyHierarchical & 30 & 2.61 & 3.89 & \textbf{4.05} \\
Barman-BDI & 20 & \textbf{16.42} & 15.40 & 15.05 \\
Blocksworld-GTOHP & 30 & \textbf{25.66} & 22.56 & 21.11 \\
Blocksworld-HPDDL & 30 & \textbf{6.74} & 0.80 & 2.91 \\
Childsnack & 30 & \textbf{27.09} & 26.78 & 21.56 \\
Depots & 30 & \textbf{25.49} & 22.05 & 24.80 \\
Elevator-Learned-ECAI-16 & 147 & \textbf{146.91} & 114.23 & 137.45 \\
Entertainment & 12 & 8.49 & 2.46 & \textbf{11.53} \\
Factories-simple & 20 & \textbf{6.12} & 3.77 & 5.99 \\
Freecell-Learned-ECAI-16 & 60 & \textbf{7.31} & 5.26 & 6.25 \\
Hiking & 30 & \textbf{24.87} & 22.07 & 18.89 \\
Lamps & 30 & 15.40 & 0 & \textbf{17.31} \\
Logistics-Learned-ECAI-16 & 80 & \textbf{60.94} & 28.48 & 49.00 \\
Minecraft-Player & 20 & \textbf{2.66} & 2.49 & 1.43 \\
Minecraft-Regular & 59 & \textbf{32.78} & 29.55 & 29.82 \\
Monroe-Fully-Observable & 20 & 18.38 & \textbf{19.09} & 11.56 \\
Monroe-Partially-Observable & 20 & 17.07 & \textbf{18.21} & 9.47 \\
Multiarm-Blocksworld & 74 & \textbf{11.66} & 1.87 & 9.06 \\
Robot & 20 & \textbf{10.92} & 10.55 & 10.75 \\
Rover-GTOHP & 30 & \textbf{21.74} & 18.05 & 18.54 \\
Satellite-GTOHP & 20 & \textbf{14.70} & 12.39 & 14.64 \\
SharpSAT & 21 & \textbf{10.21} & 8.35 & 8.61 \\
Snake & 20 & \textbf{19.79} & 19.41 & 16.57 \\
Towers & 20 & \textbf{9.51} & 8.11 & 5.98 \\
Transport & 40 & 35.05 & 31.18 & \textbf{35.49} \\
Woodworking & 30 & 26.41 & \textbf{30} & 22.01 \\
\hline
Coverage & 948 & \textbf{706} & 583 & 664 \\
Normalized coverage & 26 & \textbf{18.97} & 15.45 & 17.61 \\
IPC score &   & \textbf{604.93} & 477.0 & 529.83 \\
Normalized IPC score &   & \textbf{16.06} & 13.22 & 13.93 \\
Quality score &   & \textbf{689.53} & 520.83 & 606.49 \\
Normalized quality score &   & \textbf{18.37} & 14.17 & 15.91 \\
\end{tabular}
\centering
\caption{Performance of each planner on the IPC 2020 and IPC 2023 benchmarks. Each cell contains the IPC score for the planner in a specific domain. Total coverage, total quality score, and total IPC score, along with their normalized values per domain, are displayed in the last six rows.}
\label{table:overview_results}
\end{table}

An overview of the results for the coverage, IPC score, and quality score is presented in Table \ref{table:overview_results}. We observe that, among the 26 domains in the benchmarks, our planner achieves a better IPC score in 19 of them and also surpasses other planners in terms of coverage and quality score. Note that the quality score is computed based on the first solution DT found, and therefore does not include any plan improvement procedures, such as the one proposed by Lilotane, which allows finding the solution DT with the shortest plan within a PDT at the cost of a higher runtime. \\

\begin{figure}[h]
    \centering
    \includegraphics[width=0.5\textwidth]{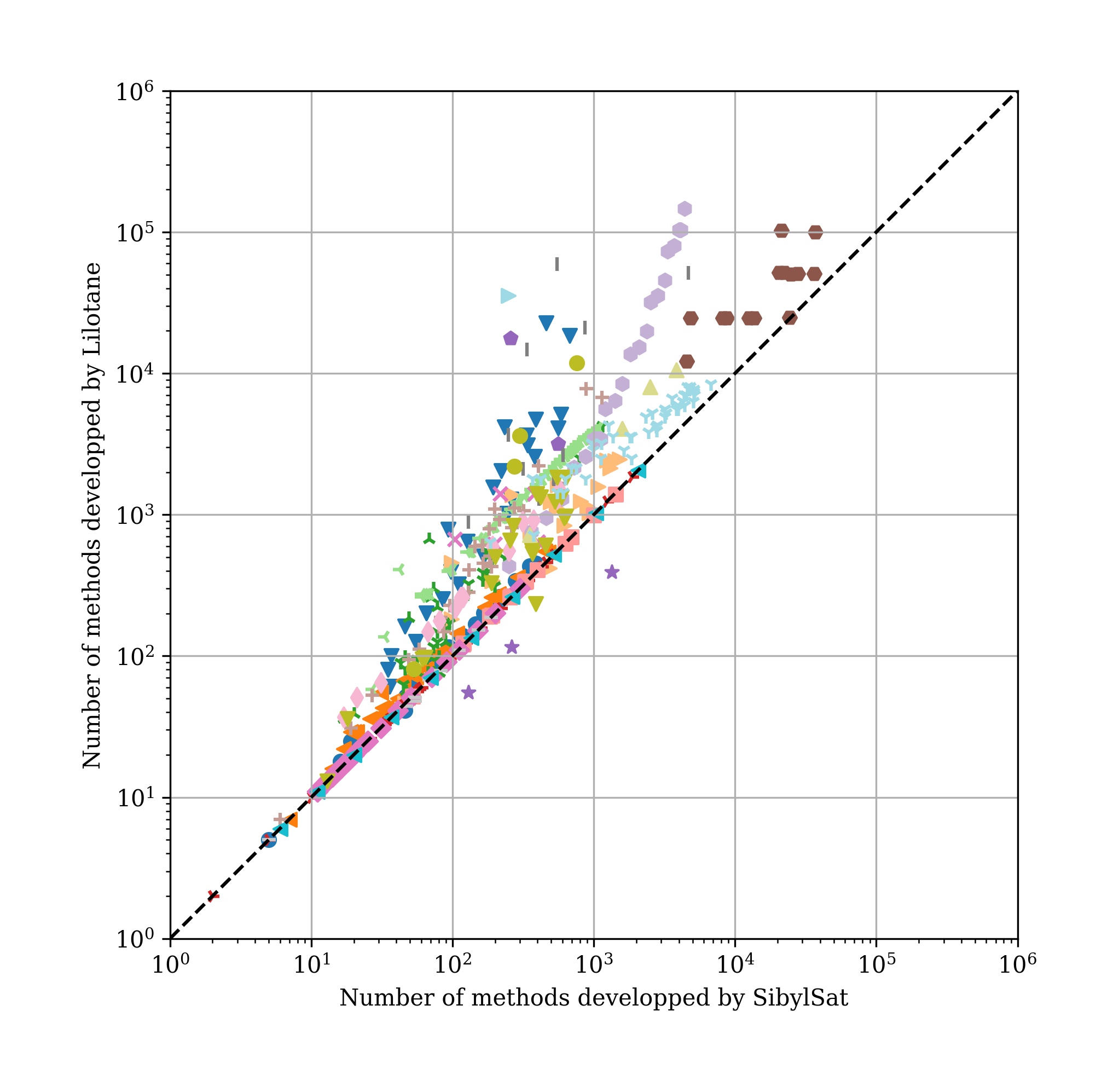}
    \caption{Comparison of the number of methods developed by SibylSat and Lilotane. A different marker is used for each benchmark.}
    \label{fig:methods_developed_comparison}
\end{figure}

To verify if the expansion strategy proposed by SibylSat effectively allows exploring a smaller part of the search space before finding a solution DT, we compare in Figure \ref{fig:methods_developed_comparison} the number of methods developed before finding a solution DT by Lilotane and SibylSat, as both use the same structure to represent the search space. The figure shows that our approach generally reduces the number of methods developed before finding a solution DT in most domains. In some domains, such as Towers and Childsnack, the number of methods developed by our planner is similar to that of Lilotane because Lilotane already develops the minimal number of nodes in its structure to find a solution DT. The only benchmark where our planner develops more methods than Lilotane is the AssemblyHierarchical domain (purple star markers in Figure \ref{fig:methods_developed_comparison}). This is reflected in the results, as it is the only benchmark where SibylSat's IPC score is lower than that of both Lilotane and PandaPIsatt-1iB. An analysis of the AssemblyHierarchical domain reveals the underlying reason for this outcome. Most of the abstract tasks in this domain are recursive and interconnected, leading to a larger set of possible effects for each task. Because of this complexity, the possible effects of each abstract task are extensive and often lead our planner to find impractical relaxation solution DT. Improving our algorithm for the inference of possible effects of the abstract tasks could likely assist with this type of problem.

\section{Related Work}

The use of SAT-based HTN planners was first introduced by Mali and Kambhampati in 1998 \cite{mali1998encoding}, although their representation of HTN problems differed significantly from the current formalisms and their encoding was incapable of handling recursive domains. Despite this early exploration, there was a two-decade gap in research that specifically focused on translating HTN planning problems into propositional logic. This changed with the introduction of totSAT in 2018 \cite{behnke2018totsat}, a SAT-based translation of TOHTN planning which outperformed other state-of-the-art HTN planners and revitalized interest in this approach. Unlike classical planning, which extends encodings iteratively along the length of the final plan, totSAT uses a breadth-first search to extend encodings along the depth of the hierarchy. Following this work, Behnke et al. refined their approach to handle partially ordered HTN planning \cite{behnke2019bringing}, optimal plan finding \cite{behnke2019finding}, and proposed techniques to prune the search space and enhance their encoding \cite{behnke2021block}.

Concurrently with totSAT, Schreiber et al. proposed a new encoding that was the first to leverage incremental SAT solving \cite{schreiber2019efficient} for HTN problems. An enhancement of this approach led to the development of the Tree-REX planner \cite{schreiber2019tree}, which explores the search space similarly to totSAT but whose translation to propositional logic is specifically designed for incremental SAT solving, providing improved performance and smaller encodings. Following this work, Schreiber et al. introduced Lilotane \cite{schreiber2021lilotane}, a successor to Tree-REX and the first lifted TOHTN SAT-based planner. Lilotane is able to avoid the costly grounding process by using a lazy instantiation approach for tasks and methods, allowing free arguments where needed, resulting in SAT formulas that are considerably smaller than previous encodings.

\section{Conclusion}

In this paper, we have presented a novel SAT-based approach for solving TOHTN planning problems that uses a SAT solver both for searching for solutions and guiding the exploration of the search space. Specifically, we demonstrated how encoding a relaxed problem can help uncover a heuristic with a SAT solver, which can then be used to identify promising areas in the search space. We have shown that our planner, which uses this heuristic to greedily explore the search space, outperforms other state-of-the-art SAT-based TOHTN planners in runtime and plan quality. This work paves the way for integrating heuristic information into SAT-based TOHTN planning.

In future work, we plan to explore methods beyond our current greedy search strategy. Specifically, we aim to determine if identifying multiple promising areas within the search space using our heuristic, then ranking and prioritizing them with classical HTN heuristics \cite{holler2018generic, holler2019guiding, holler2020htn}, could lead to improved performance.

\section{Acknowledgments}

This work has been partially supported by MIAI@Grenoble Alpes, (ANR-19-P3IA-0003).

\whileboolexpr{test {\ifnumless{\value{page}}{8}}}{
    \thispagestyle{empty} 
    \vspace*{\fill} 
    \phantom{ } 
    \clearpage
}

\bibliographystyle{unsrtnat}
\bibliography{biblio}

\end{document}